\documentclass{sig-alternate} 

\usepackage{amsmath,bbm,amssymb}
\usepackage{graphicx,verbatim}
\usepackage{hyperref}
\usepackage{url,cleveref,cite}
\usepackage{lastpage}
\usepackage{xspace}
\usepackage{subfig}
\usepackage{floatrow}
\usepackage{epsfig}
\usepackage{url}
\usepackage{multirow}
\usepackage{color}
\usepackage{algorithm}
\usepackage[noend]{algpseudocode}
\usepackage[T1]{fontenc}

\usepackage{etoolbox}
\makeatletter
\patchcmd{\maketitle}{\@copyrightspace}{}{}{}
\makeatother

\DeclareMathOperator*{\argmax}{\arg\max}
\DeclareMathOperator*{\argmin}{\arg\min}

\newcommand{\ie}{i.e.\xspace}
\newcommand{\eg}{e.g.\xspace}
\newcommand{\etc}{etc.\xspace}
\newcommand{\ignore}[1]{}

\newcommand{\reals}{\ensuremath{\mathbb{R}}}
\newcommand{\para}[1]{\smallskip \noindent {\bf #1.}}

\newcommand{\userset}{\ensuremath{\mathcal{N}}}
\newcommand{\itemset}{\ensuremath{\mathcal{M}}}
\newcommand{\pairs}{\ensuremath{\mathcal{E}}}
\newcommand{\normal}{N}

\newcommand{\ml}{Movielens\xspace}
\newcommand{\fl}{Flixster\xspace}
\newcommand{\ptv}{PTV\xspace}

\definecolor{blue_color}{rgb}{0,0,1}
\definecolor{kalamata_olive_color}{rgb}{0.47058823529411764, 0.5254901960784314, 0.4196078431372549}
  \newtheorem{thm}{Theorem}

\newlength{\figtwo}
\newlength{\figthree}
\newlength{\figfour}
\algrenewcommand\algorithmicrequire{\textbf{Input:}}
\algrenewcommand\algorithmicensure{\textbf{Output:}}
\algrenewcommand\algorithmicforall{\textbf{foreach}}

\newcounter{packednmbr}
\newenvironment{packedenumerate}{\begin{list}{\thepackednmbr.}{\usecounter{packednmbr}\setlength{\itemsep}{0pt}\addtolength{\labelwidth}{-5pt}\setlength{\leftmargin}{\labelwidth}\setlength{\listparindent}{\parindent}\setlength{\parsep}{0pt}\setlength{\topsep}{3pt}}}{\end{list}}
\newenvironment{packeditemize}{\begin{list}{$\bullet$}{\setlength{\itemsep}{0pt}\addtolength{\labelwidth}{-5pt}\setlength{\leftmargin}{\labelwidth}\setlength{\listparindent}{\parindent}\setlength{\parsep}{0pt}\setlength{\topsep}{3pt}}}{\end{list}}

\newcommand{\fbc}{\textsc{FBC}\xspace}
\newcommand{\pe}{\textsc{PointEst}\xspace}
\newcommand{\incfbc}{\textsc{IncFBC}\xspace}
\newcommand{\maxgap}{\textsc{MaxGap}\xspace}

\newcommand{\numn}{\ensuremath{n}\xspace}
\newcommand{\numm}{\ensuremath{m}\xspace}
\newcommand{\type}{\ensuremath{t}\xspace}
\newcommand{\typeset}{\ensuremath{\mathcal{T}}}
\newcommand{\bias}{\ensuremath{z}\xspace}
\newcommand{\Bias}{Z}
\newcommand{\diff}{\delta}
\newcommand{\expect}{\mathbb{E}}
\newcommand{\score}{L}
\newcommand{\sign}{\ensuremath{\mathop{\mathrm{sgn}}}}

\makeatletter
\def\@copyrightspace{\relax}
\makeatother

\newcommand{\techrep}[2]{#1}
\begin{document}

\title{Recommending with an Agenda: Active Learning of Private Attributes using Matrix Factorization}
\author{{ Smriti Bhagat, Udi Weinsberg, Stratis Ioannidis, Nina Taft} \\ 
\affaddr{Technicolor, Los Altos, CA }\\ 
\email{\small\{smriti.bhagat, udi.weinsberg, stratis.ioannidis, nina.taft\}@technicolor.com}\\
}
\maketitle

\makeatletter{}\begin{abstract}
Recommender systems leverage user demographic information, such as age, gender, etc., to personalize recommendations and better place their targeted ads. Oftentimes, users do not  volunteer this information due to privacy concerns, or due to a lack of initiative in filling out their online profiles. We illustrate a new threat in which a recommender learns private attributes of users who do not voluntarily disclose them. We design both  passive and active attacks that solicit ratings for strategically selected items, and could thus be used by a recommender system to pursue this hidden agenda.
Our methods are based on a novel usage of Bayesian matrix factorization in an active learning setting. Evaluations on multiple datasets illustrate that such attacks are indeed feasible and use significantly fewer rated items than  static inference methods. Importantly, they succeed without sacrificing the quality of recommendations to users.

\ignore{
In this work, we present a new attack that a recommender
system could use to pursue a hidden agenda of inferring
private attributes for users that do not voluntarily
disclose them. The recommender can exploit the usual user rating process to solicit ratings for specific items (strategically selected) in order to learn the private attribute and bypass the user's attempt at privacy. In this work, we propose a method for inferring a user's demographic attributes in an active learning setting, where the recommender system strategically solicits user ratings to improve the accuracy of inference of the user's private attributes. Our method is based on Bayesian matrix factorization, which has been extensively --and successfully -- used  in the past for rating prediction. Our evaluations show that by carefully selecting the items, we can obtain similar or better accuracy of inferring demographic information  with a much smaller number of rated items as compared with using static inference techniques, while preserving the quality of recommendations.
}
\end{abstract}

\makeatletter{}\section{Introduction}

Recommender systems rely on knowing their users -- not just their preferences (\ie, ratings
on items), but also their social and demographic information, \eg, age, gender, political affiliation, and ethnicity \cite{KK2013,hotel2012}. A rich user profile allows a recommender system to better personalize its services, and at the same time enables additional monetization opportunities, such as targeted advertising.

\ignore{Recommender systems thrive on knowing their users -- not just their preferences (\ie, ratings
on items), but also their demographics (\eg, age, gender, ethnicity, and location).
These can
be used not only for improving the recommendations offered to users, but also to enable
additional services, such as targeted advertising.
}

\ignore{
Several applications such as online advertising, personalized content delivery (news, blogs, \emph{etc}.) and product recommendations benefit greatly from demographic profiling of users. With the ability to track a user's online activity, such as the websites the user browses, the movies she watches online, and the products she buys, there is a potential to perform fine-grained user profiling that is well known to be useful for online advertising.
For instance, if a system determines that a user searching for handbags and jewelry is likely to be a female, the system can use this information to target ads for makeup to this user.
}

Users of a recommender system know they are disclosing their preferences (or ratings) for movies, books, or other items (we use movies as our running example). 
A recommender may obtain additional social and demographic information about its users by explicitly soliciting it \cite{KK2013, hotel2012}. While some users may willingly disclose such information, others may be more privacy-sensitive and elect to only reveal their ratings. Privacy research has shown that some users are uncomfortable revealing their demographic data to personalization systems \cite{paradox2006, clientside-chi2014}. 
Even when such services provide transparency about their data collection and use practices \cite{paradox2006}, some users are unwilling to disclose personal data despite the allure of personalized services.
In \cite{KK2013} the authors conduct a small scale user study on Amazon Turk that examines how to motivate users to disclose their demographic data.

For users who wish to withhold some demographic information,  a recommender can still undermine their attempt at privacy. In  previous work~\cite{blurme:2012}, we show that users' movie ratings can be used to predict their gender with 80\% accuracy. Other studies  also show the potential to infer demographics from a range of online user activities~\cite{FBlikes2013,Bhagat:2007,Mislove:2010,Bi-www2013}. 
In this work, we consider a recommender system that offers a legitimate service, yet is simultaneously malicious:  it purposefully attempts to extract certain attributes from users who choose to withhold them. Unlike previous work that studies static attacks on the complete data, we consider an \emph{active learning} setting, in which the recommender system aims to efficiently (quickly and accurately) infer a user's private attribute via interactive questioning. Recommender systems routinely 
ask users to rate a few items, as a means to address a ``cold start'' setting, or to improve the quality of recommendations. We leverage these instances of interactions with the user, alongside with the observation that item selection is at the recommender's discretion, to propose a new attack.
 We hypothesize that if the
sequence of questions (items to rate) is carefully selected, the recommender system can quickly (so as not to be detected by the user) determine a user's private attribute with high confidence, thereby violating her privacy. 
A key idea in the design of this attack is to leverage matrix factorization (MF) as the basis for inference.
Most recommender systems use matrix factorization (MF) models
as a building block for providing recommendations~\cite{Koren:2009}. While MF is well understood for rating prediction, it has generally not been applied for inference. To the best of our knowledge, this
paper is the first to leverage MF as the basis for building both (a) an inference method of private attributes using item ratings and (b) an active learning method that selects items in a way that maximizes inference confidence in the smallest number of questions.

\ignore{
In this paper, we study the question of how well private binary attributes can be learned simply via a set of ratings.
We formulate the problem of inferring a private binary attribute as an \emph{active learning} task. To the best of our knowledge, we are the first to address this problem and design a solution. Crucially, we propose an approach that leverages Matrix factorization (MF) as the basis for demographic inference. While MF is well understood for ratings prediction, it has generally not been applied for inference, let alone in an active learning setting. Moreover, while MF has been widely validated for profiling items and predicted ratings, it not a priori clear how to leverage it for inference of demographic information.
}

Our contributions are as follows:
\vspace{-1mm}
\begin{packeditemize}
\item First, we propose a novel classification method for determining a user's binary private attribute -- her \emph{type} -- based upon ratings alone.  In particular, we use matrix factorization to learn item profiles and type-dependent biases, and show how to incorporate this information into a  classification algorithm. This classification is consistent with Bayesian  matrix factorization. \item Second, we demonstrate that the resulting classification method is well suited for learning of a user's type. A simple \emph{passive} approach, ordering items based on a set of weights computed off-line, works quite well in many cases. Beyond this, we design an \emph{active learning} algorithm for selecting the next item to ask a user to rate:  each  selection 
maximizes the expected confidence of the private attribute's inference. Equivalently,  the selections of the active learning algorithm minimize the expected risk of misclassifying the user's private attribute.
\item Third, we show that our active learning method is very efficient, as item selection can reuse computations made during previous selections. We show that this reduces the na\"ive solution that is cubic in the number of ratings, to one that is quadratic in the number of ratings.  \item Fourth, we extensively evaluate the above classifier and selection methods on three real-world datasets: \ml, \fl and Politics-and-TV.  We show that our methods consistently outperform other baselines; with only 10 questions, we achieve between 3-20\% higher classification accuracy on different datasets. Importantly, such an attack can be carried out without any sacrifice to the recommendations made to the user.
\end{packeditemize}

\techrep{In the remainder of the paper, we review related work (\Cref{sec:related}) and formulate our problem~(\Cref{sec:sysdesc}). We then present our classifier (\Cref{sec:fbc}), our active learning method (\Cref{sec:strategy}), and our empirical results (\Cref{sec:eval}).}{}

\makeatletter{}\section{Related Work}
\label{sec:related} 

A number of studies have shown that user demographics can be inferred from various types of online user activity. For example, Bhagat \emph{et al.}~\cite{Bhagat:2007} show that it is possible to learn age and gender information from blogs. Mislove \emph{et al.}~\cite{Mislove:2010} study data from online social networks and illustrate that even when only a fraction of users provide profile attributes (such as location, interests, \etc), it is possible to infer these  attributes among users who do not disclose them.  Bi~\emph{et al.}~\cite{Bi-www2013} show how demographics can be inferred from search queries, and  Kosinski \emph{et al}.~\cite{FBlikes2013} show that several personality traits, including political views, sexual orientation, and drug use can be accurately predicted from  Facebook ``likes''.

Recommender systems were shown to be exploitable by several works utilizing off-line attacks \cite{Arvind11,Narayanan:2008,blurme:2012}. Closest to our setting, Weinsberg et al.~\cite{blurme:2012} empirically studied how to infer a user's gender from her movie ratings  using a variety of different classifiers, showing that logistic regression and SVMs succeed   with an accuracy close to 80\%. We depart from~\cite{blurme:2012} in multiple ways. First, we introduce a novel factor-based classifier, that relies on the Bayesian assumptions behind MF. Second, we study a recommender system in an adversarial setting that actively adapts item selection to quickly learn the private attributes. Finally, we establish that our  classifier is very well suited for this task.

The  Bayesian model underlying MF (discussed in detail in \Cref{sec:model}) was recently employed by Silva and Carin~\cite{silva2012active} to actively learn the actual factors (\emph{i.e.}, the user and item profiles) in MF. More specifically, the authors consider a recommender system that adaptively selects which items to ask its users to rate in order to diminish the entropy of its user and item profiles as quickly as possible. The entropy estimation is based on the Gaussian noise and prior assumptions underlying MF, which we also employ in our work. A variety of  active learing objective were also studied by Sutherland~et al.~\cite{sutherland2013active}, including minimizing the prediction error on unrated items, reducing the profile uncertainty, and identifying highly rated items. We depart from the above works as the goal of our learning task is to discover a user's demographic information, captured by a categorical type, rather the above objectives motivated by rating prediction. 
\techrep{
Biases have been used extensively to improve prediction performance in MF \cite{Koren:2008,Koren:2009}. Our introduction of demographic-specific biases is not for improving prediction per se--though this does happen; rather, incorporating such biases allows us to use the classic MF model as a basis for classification and, subsequently, active learning.}{}

In the classic active learning setting~\cite{dasgupta2005analysis,golovin2010adaptive}, a learner wishes to disambiguate among a set of several possible hypotheses, each represented as a function over a set of inputs. Only one of the hypotheses is valid; to discover it, the learner has access to an  oracle that returns the evaluation of the valid hypothesis on a given input. In the case of a \emph{noiseless} oracle, that always returns the correct evaluation on a query, Generalized Binary Search (GBS) discovers the valid hypothesis in a number of queries within a polylogarithmic factor from the optimal~\cite{dasgupta2005analysis,golovin2010adaptive}.
Our setup can be cast into the above framework in the context of a \emph{noisy} oracle, whose evaluations may not necessarily be exact. GBS is known to yield arbitrarily suboptimal results in the presence of noise~\cite{golovin2010near}. Though algorithms for restricted noise models exist (see, {e.g.},~\cite{nowak2009geometry} and~\cite{golovin2010near}), no algorithm with provable performance guarantees is known in the presence of an oracle with arbitrary noise. Unfortunately, none of the existing models apply to the noisy setting we encounter here.

\makeatletter{}\section{System Description}\label{sec:sysdesc}
\subsection{Problem Statement}

\techrep{
\begin{figure}[!tb]
{\centering \includegraphics[width=0.6\figtwo]{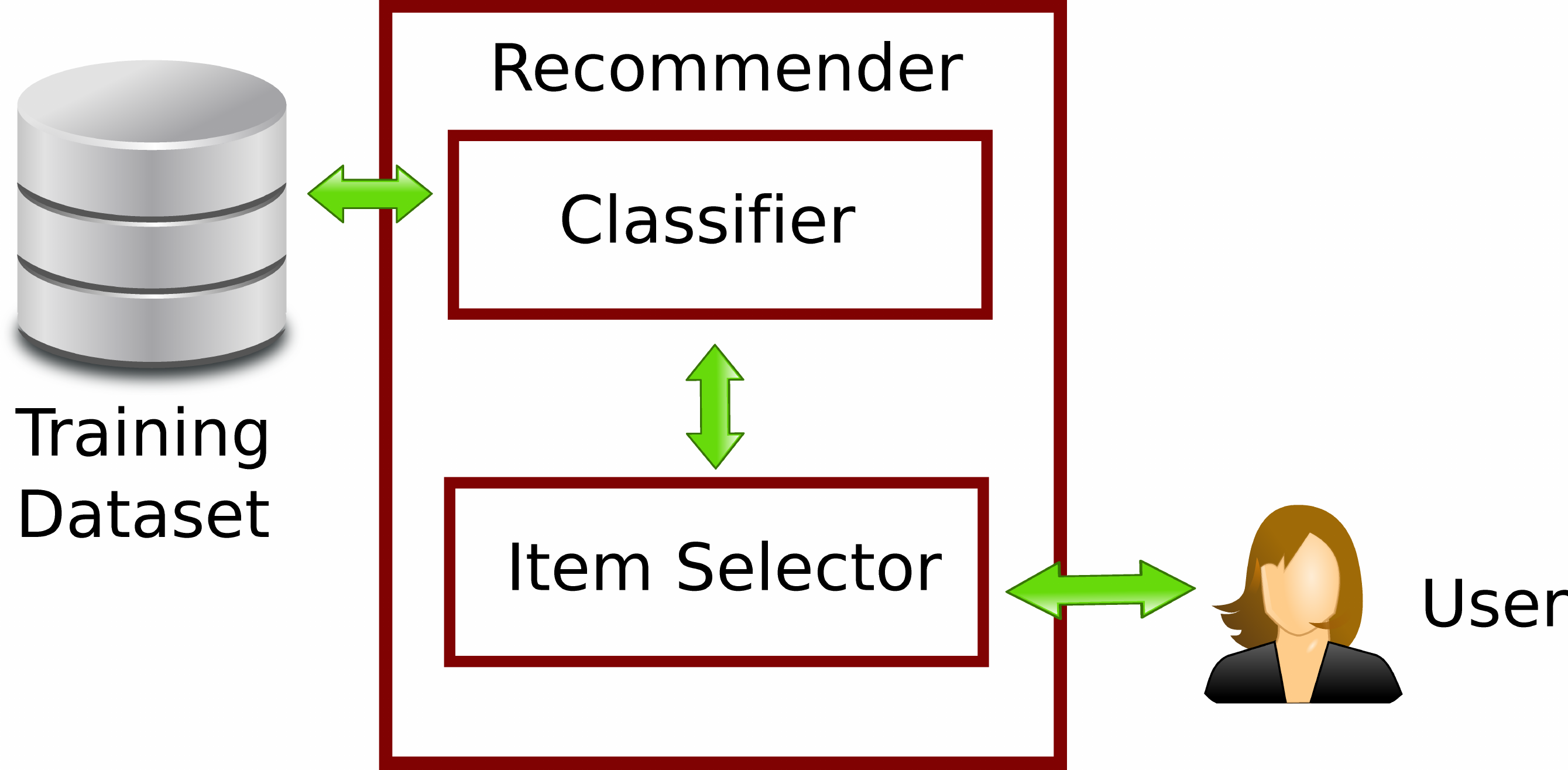}}
\caption{System description. The recommender system uses a dataset of user ratings to train a type classifier. An item selection process proposes items to the user, which she subsequently rates; these ratings are used to infer her type.}\label{systemfig}
\end{figure}
}{}
 
We consider a recommender system\techrep{, depicted in \Cref{systemfig},}{}
 that provides a legitimate item recommendation service, but at the same time maliciously seeks to infer a private user attribute. The system has access to a dataset, provided by non-privacy-sensitive users, that contains item ratings as well as a categorical variable, which we refer to as the user \emph{type}. The type is a private attribute such as gender, age, political affiliation, \etc
A new user, who is privacy sensitive (\ie, her type is unknown) interacts with the system. The recommender system actively presents items for the user to rate, masquerading it as a way to improve recommendations 
in the cold-start setting. 
In this context, our goal is twofold:
\begin{packedenumerate}
\item We wish to design a \emph{type classifier} that discovers the type of the user based on her ratings. We seek to leverage the  latent factor model prevalent in matrix-factorization, a technique successfully used for rating prediction by recommender systems.
\item We wish to address the problem of \emph{actively learning} a user's type. We aim to design an item selection method, that determines the order in which items are shown to a user for her to rate. The best order finds the user's type as quickly as possible.
\end{packedenumerate}
For the attack to be considered successful, the recommender system needs  to obtain high confidence in the value of the inferred type, with a minimum number of questions posed to the user. As our classifier and item selection methods rely heavily on matrix factorization, we review this as well as the latent factor model that underlies it below.

\makeatletter{}\subsection{Data Model \& Matrix Factorization}
\label{sec:model}

 We use the following notation to describe the training dataset of the recommender. The dataset comprises of ratings to $\numm$ items in set $\itemset\equiv\{1,\ldots,\numm\}$ given by $\numn$ users in set $\userset\equiv\{1,\ldots,\numn\}$. We denote by $r_{ij}$ the rating of user $i\in \userset$  to item $j\in \itemset$, and  by $ \pairs\subset \userset\times \itemset$ the set of user-item pairs $(i,j)$, for which a rating $r_{ij}$ is present in the dataset. Each user  is characterized by a categorical type, which captures demographic information such as gender, occupation, income category, \emph{etc.} Focusing on binary types, we denote by $\type_i\in\typeset \equiv\{+1,-1\}$ the type of user $i\in \itemset$.

We assume that the ratings are generated from the standard generative model used in matrix factorization, augmented with type-dependent biases. More specifically,  there exist latent factors $u_i\in \reals^d$, $i\in \userset$, and $v_j\in\reals^d$, $j\in \itemset$ (the \emph{user} and \emph{item profiles}, resp.) such that ratings are:\begin{align}
r_{ij} = u_i^Tv_j+\bias_{j\type_i}+\epsilon_{ij}, \quad (i,j)\in \pairs\label{bilinear}
\end{align}
where $\epsilon_{ij}\sim \normal(0,\sigma_0^2)$ are independent Gaussian noise variables and $\bias_{jt}$ is a \emph{type bias}, capturing the effect of a type on the item rating. Our model is thus parametrized by  $U=[u_i^T]_{i\in \userset}\in \reals^{\numn\times d}$,$V=[v_j^T]_{j\in \itemset}\in \reals^{\numm\times d}$, and $\Bias=[\bias_{j,t}]_{j\in \itemset,t\in\typeset}\in\reals^{\numm\times|\typeset|}$.
We further assume a \emph{prior} on user and item profiles:  for all $i\in \userset$, $j\in\itemset$,
\begin{align}
u_i \sim \normal(\mathbf{0},\sigma_u^2I),\text{ and }  v_j\sim \normal(\mathbf{0},\sigma_v^2 I), \label{priors}\end{align}
\emph{i.e.}, profiles are sampled from independent zero-mean multivariate Gaussian priors.

The Gaussian  priors \eqref{priors} are used in many works on so-called Bayesian matrix factorization (see,  \emph{e.g.},~\cite{salakhutdinov2008probabilistic,nakajima2010implicit,silva2012active}). 
Under \eqref{bilinear} and \eqref{priors}, the maximum likelihood estimation of the model parameters reduces to the standard \cite{Koren:2008,Koren:2009}
 minimization of the  (non-convex) regularized error:\footnote{Note that, as is common practice, to ensure that the profiles $U,V$ obtained  by  \eqref{mle} are invariant to a translation (shift) of the ratings, we do not regularize the category biases (or, equivalently, we assume no prior on $\Bias$).
}
\begin{align}
\begin{split}
\min_{U,V,\Bias}\!\!\sum_{(i,j)\in \pairs}\!\! (r_{ij}&\!-\!u_i^Tv_j\!-\!\bias_{j\type_i})^2\!+\!\lambda \!\sum_{i\in \userset}\!\|u_i\|_2^2\!+\!\mu \!\sum_{i\in \itemset}\!\!\|v_j \|_2^2
\end{split}\label{mle}
\end{align}
with $\lambda=\frac{\sigma_0^2}{\sigma_u^2}$ and $\mu=\frac{\sigma_0^2}{\sigma_v^2}$.
Given a dataset of ratings $r_{ij}$, $(i,j)\in \pairs$ and types $\type_i$, $i\in\userset$, the parameters $U,V,\Bias$ can be computed as local minima to \eqref{mle} through standard methods \cite{Koren:2009}, such as gradient descent or alternating minimization, while $\lambda$ and $\mu$ are computed through cross-validation.

\section{A Factor-Based Classifier}\label{sec:fbc}
We now turn our attention to the following classification problem. Suppose that the recommender system, with access to the dataset of ratings and types, has computed a set of item profiles $V$ as well as a set of biases $\Bias$, \emph{e.g.}, by minimizing \eqref{mle} through gradient descent. A new user arrives in the system and submits ratings for items in some set $A\subseteq \itemset$, \emph{but does not submit her type.} In order to bypass the user's attempt at privacy, we need to construct a classifier to infer the type of this new user.

In this section, we present a classifier that  uses the item profiles and biases (\emph{i.e.}, the latent factors obtained through matrix factorization) to accomplish this task. We refer to this classifier as a \emph{Factor-Based Classifier}~(\fbc). Crucially,  \fbc is consistent with the Bayesian model of matrix factorization presented in the previous section. In particular, it amounts to the maximum a-posteriori estimation of the type under the bi-linear noise model \eqref{bilinear} and the priors \eqref{priors}.

\subsection{Type Posterior}
For $A\subset\itemset$ the set of items for which the user submits ratings, we introduce the following notation.
We denote by $r_A\equiv [r_j]_{j\in A}\in \reals^{|A|}$ the vector of ratings provided by the user,
by $V_A \equiv [v_j^T]_{j\in A}\in\reals^{|A|\times d}$ the matrix of profiles for items rated, and by
 $\bias_{A\type} \equiv [\bias_{j\type}]_{j\in A}\in \reals^{|A|}$ the vector of type-$\type$ biases for items rated.

As in the previous section, we assume the new user has an unknown profile $u\in \reals^d$ and a type $\type\in\{-1,+1\}$, such that the ratings she submits follow \eqref{bilinear}, \emph{i.e.},
\begin{align}
r_j = u^Tv_j+\bias_{j\type}+\epsilon_j, \quad j\in A,  \label{linear}
\end{align}
where $\epsilon_j\sim \normal(0,\sigma_0^2)$.
That is, conditioned on $u$ and $\type$, the ratings $r_A=[r_j]_{j\in A}\in \reals^{|A|}$ given to items in $A\subset [M]$ are distributed as follows:
\begin{align}
\Pr(r_A\mid u,\type)  ={e^{-\|r_A -  V_Au - z_{A\type}\|^2_2/2\sigma_0^2} }/{\big(\sigma_0 \sqrt{2\pi}\big)^{|A|}}\label{lineardistr}
\end{align}
where $\sigma_0^2$ is the noise variance.

Moreover, we assume as in the previous section that profile $u$ follows a zero-mean Gaussian prior with covariance $\sigma_u^2I$, and that the type follows a uniform prior (\emph{i.e.}, each of the two types is equally likely), i.e.: \begin{align}\Pr(u,t) =  0.5 e^{-\|u\|_2^2/2\sigma_u^2} /\big(\sigma_u \sqrt{2\pi }\big)^d\label{prior} \end{align}

\subsection{Classification}
Under the above assumptions, it is natural to  classify the incoming user using maximum a posteriori estimation of the type $\type$. In particular, \fbc amounts to
\begin{align}\hat{\type}(r_A) =\textstyle\argmax_{\type \in \typeset}\Pr(\type\mid r_A).\label{maxapost} \end{align}
Under this notation, \fbc can be determined as follows:
\begin{thm}\label{thm:fbc} Under noise model \eqref{lineardistr} and prior \eqref{prior}, the \fbc classifier  is given by
\begin{align}\label{fbc}
\hat{\type}(r_A) = \sign(\diff_A^T M_A \bar{r}_A)
\end{align}
where
$\bar{r}_A \equiv r_A - \frac{\bias_{A+}+\bias_{A-}}{2},$
$\diff_A \equiv \frac{\bias_{A+}-\bias_{A-}}{2}$,  $M_A\equiv I-V_A\Sigma_A^{-1} V_A^T$, and
$\Sigma_A \equiv \lambda I + V_A^TV_A $, for $\lambda=\frac{\sigma^2_0}{\sigma^2_u}$.
\end{thm}
\techrep{
\begin{proof}
Under model \eqref{lineardistr} and prior \eqref{prior}, conditioned on type $\type$,
the ratings $r_A$ a user gives items in a set $A\subseteq[M]$ are distributed according to:
\begin{align} \Pr(r_A\mid \type)  &=\frac{ e^{\frac{( r_A-\bias_{A\type})^T\left(V_A\Sigma_A^{-1}V_A^T-I\right)( r_A-\bias_{A\type})}{2\sigma_0^2}}
}{(\sigma_0 \sqrt{2\pi})^{|A|}(\sigma_u/\sigma_0)^d \sqrt{\det(\Sigma_A)}
}
  \label{uncond}
\end{align}
where $\Sigma_A \equiv \lambda I + V_A^TV_A$ and $\lambda\equiv \frac{\sigma_0^2}{\sigma_u^2}$.
Hence, the posterior probability of the user's type is given by:
\begin{align}
\Pr(\type\mid r_A) &=  \frac{e^{(r_A\!-\!z_{At})^T\left(V_A\Sigma_A^{-1}V_A^T\!-\!I\right)(r_A\!-\!z_{At})/2\sigma_0^2}}{\displaystyle\sum_{\type'\in \typeset}\!\!\!
e^{(r_A\!-\!z_{At'})^T\left(V_A\Sigma_A^{-1}V_A^T\!-\!I\right)(r_A\!-\!z_{At'})/2\sigma_0^2}}
\label{gender}
\end{align}
Type $\type=+1$ is thus most likely if
\begin{align*}
\begin{split}
(r_A-z_{A+})^T&\left(V_A\Sigma_A^{-1}V_A^T-I\right)(r_A-z_{A+})-\\
& (r_A-z_{A-})^T\left(V_A\Sigma_A^{-1}V_A^T-I\right)(r_A-z_{A-})\geq 0
\end{split}
\end{align*}
and it is easy to verify that \eqref{fbc} follows.\end{proof}}
{The proof can be found in the extended version of our paper~\cite{techrep}.} There are several important observations to be made regarding \fbc, as defined by Theorem~\ref{thm:fbc}.

\para{Set of Classifiers} We first note that \fbc in fact defines a  \emph{set} of classifiers, each parametrized by set $A\subseteq \itemset$:  each such classifier $\hat{\type}:\reals^{|A|}\to \{-1,+1\}$ takes as input any possible set of ratings $r_A\in \reals^{|A|}$ as input. Note however that all classifiers are trained \emph{jointly} from the ratings dataset: this ``training'' amounts to  determining the item profiles $V$ and the item biases $Z$ through matrix factorization. With $V$ and $Z$ learned, when presented with ratings $r_A$, the classifier can compute the vectors $\bar{r}_A$, $\delta_A$ and the matrix $M_A$ needed to determine the type. Indeed, the fact that training the classifier amounts to computing the latent factors/item profiles is consistent with the observation that \fbc shares the same underlying Bayesian assumptions as matrix factorization.

\techrep{
\para{Relationship to LDA} Second, for a given set of items $A$, the classifier defined over ratings $r_A$ is a \emph{linear} classifier. In particular, \eqref{fbc} defines a hyperplane in $\reals^{|A|}$ above which the user type is classified as $+1$ and below which the type is classified as $-1$. In fact, when restricted to a specific set of items $A$, \eqref{fbc} can be seen as classification through Linear Discriminant Analysis (LDA) \cite{hastie}. More formally, the proof of Theorem~\ref{thm:fbc} \techrep{}{in \cite{techrep}} uses the fact that the ratings $r_A\in \reals^{|A|}$ are normally distributed with a mean that depends on the user type and a covariance $M_A=(I-V_A\Sigma_A^{-1}V_A)$, as defined in Theorem~\ref{thm:fbc}. As such, given a uniform prior on the types, the most likely type can indeed be determined through LDA, which yields a decision boundary precisely as in \eqref{fbc} (see, \emph{e.g.}, eq.~(4.9) pp.~108 of \cite{hastie}). Nevertheless, \fbc significantly departs from classical LDA in that all classifiers across all sets $A\subseteq \itemset$ are trained jointly.
}{}

\techrep{
\para{Type Priors and Multi-Category Types} A similar analysis to the one we discussed below allows us to extend our results to non-uniform type priors, as well as to the non-binary case. In the case of non-uniform type priors, the decision boundary of the classifier has an additional term, yielding:
$$\diff_A^T M_A \bar{r}_A + \log\frac{\pi_+}{\pi_-}\geq 0$$
where $\pi_\type$ the prior probability of each type. Finally, multi-category classification can be reduced to binary classification, by comparing all possible pairs of types through \eqref{fbc}, and selecting the type that dominates all other types. Equivalently, the decision boundaries of each type tesselate $\reals^{|A|}$, uniquely mapping each vector $r_{A}$ to the most likely type.
}{}

\makeatletter{}\section{Learning Strategies}\label{sec:strategy}

The second task in designing this threat is to find a user's type {\em quickly}. In what follows, we present two strategies for addressing this problem. The first is a \emph{passive} strategy: the recommender presents items to the user in a predetermined order, computed off-line. The second is an \emph{active} strategy: the recommender selects which item to present to the user next based on the answers she has given so far. Both strategies are extensively evaluated in \Cref{sec:eval}.  

\ignore{
The active learning algorithm we propose selects an item at each step whose ratings minimize the expected risk of \fbc (or, equivalently, maximize its confidence). It has several important advantages. First, the expected risk of \fbc can be computed exactly in closed form; compared to heuristic ``point-estimates'' of risk, this allows to incorporate not only the effect that a rating has on the performance of the classifier, but also how variable our estimate of a rate is -- we discuss this in more detail in~\Cref{sec:pointest}.
Second, we demonstrate how this computation can be performed incrementally, thus significantly reducing the computational complexity of item selection operations.
}
\subsection{\maxgap: A Passive Strategy}
A simple, passive method for presenting items to the user is to (a) sort items $j \in \mathcal{M}$ with respect to $|\delta_j|$, the absolute value of the gap between the type biases, and (b) present items to the user in decreasing order. We call this strategy \maxgap: intuitively, this method identifies the most discriminative items in the dataset, and solicits responses to these items first. 
Clearly, \maxgap does not take into account (or adapt to) how the user rates the items presented so far. Despite this limitation,
 as we will see in  \Cref{sec:eval}, this simple strategy actually performs surprisingly well in many cases, especially when there exist many highly discriminative items. When this is not the case, however, an active strategy is needed, motivating our second method.

\subsection{\protect\textsc{FBC-Selection}: An Active Strategy}
Our active method, \fbc-\textsc{Selection}, 
is summarized in Algorithm~\ref{algo:fbc}.
 Let $\hat{\type}$ be the \fbc classifier defined by \eqref{fbc}. Given observed ratings $r_A\equiv [r_j]_{j\in A}\in \reals^{|A|}$, for some $A\subset \itemset$, we define the \emph{risk} $L(\hat{\type}(r_A))$  of the classifier to be 0 if the prediction is correct, and 1 otherwise. Conditioned on $r_A$, the expected risk is
$\expect[L(\hat{t}(r_A))\mid r_A]=1-\Pr(\hat{t}(r_A)\mid r_A), $
\emph{i.e.},  it equals the 1 minus the \emph{confidence} of the classifier,
the posterior probability of the predicted type, conditioned on the observed ratings. Since, by \eqref{maxapost}, \fbc selects the type that has the maximum posterior probability, the expected risk is at most (and the confidence at least) 0.5.
\begin{algorithm}[!t]
  \caption{\fbc-\textsc{Selection}}\label{algo:fbc}
{\fontsize{7}{7}\selectfont
\begin{algorithmic}[1]
    \Require{ Item profiles $V$, item biases $Z$, confidence $\tau$}
    \State $A \leftarrow \emptyset$
    \State $r_A \leftarrow \emptyset$
    \Repeat
	\For{ \textbf{all} $j\in \itemset\setminus A$}
	 \State   Compute  $\score_j $ through \eqref{eqn:risk}
	\EndFor
         \State $j^* \leftarrow\displaystyle \argmin_{j\in \itemset\setminus A} \score_j$
	\State Query user to obtain $r_{j^*}$
        \State $A\leftarrow A\cup \{j^*\}$, $r_{A}\leftarrow r_A\cup r_{j^*}$
     \Until{ $\Pr(\hat{t}(r_A)\mid r_A)>\tau$}
  \end{algorithmic}
}
\end{algorithm}

\fbc-\textsc{Selection} proceeds greedily, showing the item that minimizes the classifier's expected risk at each step. More specifically, let $A$ be the set of items whose ratings have been observed so far. To select the next item to present to the user, the algorithm computes for each item  $j\in\itemset\setminus A$, the expected risk $\expect[L(\hat{t}(r_A\cup r_j))\mid r_A ]$ if rating $r_j$ is revealed:
$$\textstyle\int_{r_j\in \reals}(1- \Pr(\hat{t}(r_{A}\cup r_j)) \mid r_{A}\cup r_j  )\Pr(r_A\cup r_j\mid r_A)dr_j.$$
This expected risk depends on the \emph{distribution} of the unseen rating $r_j$ \emph{conditioned} on the ratings observed so far.

Under noise model \eqref{lineardistr} and prior \eqref{prior}, the expected risk for each item $j$ can be computed in a closed form. In particular, let $M_A$, $\bar{r}_A$, $\diff_A$ be as defined in Theorem~\ref{thm:fbc}. Then,  the expected risk when revealing the rating of item $j$ is proportional to the following quantity, derived in \techrep{the appendix:}{\cite{techrep}:}
\begin{align}
L_j\! =\! \!\!{\int_{r_j}\!\!\!\!\! e^{-\frac{\bar{r}_{A_j}^T \!M_{A_j}\!\bar{r}_{A_j}\!+ \!2|\diff_{A_j}^T\!M_{A_j} \!\bar{r}_{A_j}\!|\!+\!\diff_{A_j}^T\! M_{A_j}\!\diff_{A_j}}{2\sigma_0^2}}\!\!dr_j
}/\!{\sqrt{\det{\!\Sigma_{A_j}}}}\!\!\!\label{eqn:risk}\end{align}
where $A_j=A\cup \{j\}$. The integration above is w.r.t.~$r_j$, \emph{i.e.}, the predicted rating for item $j$. The outcome of the above integration can be computed in closed form in terms of the error function $\mathrm{erf}$ (i.e., \emph{no numerical integration is necessary}).  
The  formula can be found \techrep{in the appendix.}{in~\cite{techrep}.} 
 Each iteration amounts to computing the ``scores'' $L_j$ for each item $j$ not selected so far, and picking the item with the lowest score (corresponding to minimum expected risk). Once the item is presented to the user, the user rates it, adding one more rating to the set of observed ratings. The process is repeated until the confidence of the classifier (or, equivalently, the expected risk)  reaches a satisfactory level.

\subsection{\protect\textsc{IncFBC}: An Efficient Implementation}
\label{efficientimpl}
\textsc{FBC-Selection} requires the computation of the scores $L_j$ after each interaction with the user. Each such calculation involves computing the determinant $\det( \Sigma_{A_j})$, as well as the matrix $M_{Aj}=(I-V_{A_j}\Sigma^{-1}_{A_j}V_{A_j}^T)$, both of which appear in \eqref{eqn:risk}. Though having a closed form formula for \eqref{eqn:risk} avoids the need for integration, computing each of these matrices directly from their definition involves a considerable computational cost.
In particular, the cost of computing $\Sigma_A=\lambda I +V^T_AV_A$ is $O(d^2|A|)$. Computing $\Sigma_A^{-1}$ and $\det( \Sigma_{A_j})$ have a cost $O(d^3)$ multiplications using, \emph{e.g.}, LU-decomposition, which can be dropped to $O(d^{2.807})$ using Strassen's algorithm for multiplication \cite{cormen2001introduction}. Finally, the computation of $M_A$ requires $O(|A|\times d^2+|A|^2\times d)$ multiplications. As a result, the overall complexity of computing $L_j$ directly is $O(|A|\times d^2+|A|^2\times d +d^{2.807})$.
However, the performance of these computations can be significantly reduced by constructing these matrices incrementally: $M_{A_j}$, $\Sigma_{A_j}^{-1}$ and $\det(\Sigma_{A_j})$ can be computed efficiently from $M_{A}$, $\Sigma_{A}^{-1}$, and $\det(\Sigma_A)$, exploiting the fact that $\Sigma_{A_j}=\Sigma_A+v_jv_j^T,$ \emph{i.e.}, it results from $\Sigma_i$ through a \emph{rank-one} update. We discuss this below.

\para{Incremental computation of $\det(\Sigma_{A_j})$} The determinant can be computed incrementally using only  $O(d^2)$ multiplications through the Matrix Determinant Lemma \cite{matdet}, namely:  \begin{align}
\det(\Sigma_{A_j}) = (1+v_j^T\Sigma_A v_j)\det(\Sigma_A).\label{matdetform}
\end{align}

\para{Incremental computation of $\Sigma_{A_j}^{-1}$} The inverse of a rank-one update of a matrix can be computed through the Sherman-Morisson formula \cite{sherman-morrison}, which gives:
\begin{align}
\Sigma_{A_j}^{-1} = \Sigma_A^{-1} - {\Sigma_A^{-1}v_jv_j^T\Sigma_A^{-1}}
              /{(1+v_j^T\Sigma_A^{-1}v_j)},\label{sm}
\end{align}
and again reduces the number of multiplications to $O(d^2)$.

\para{Incremental computation of $M_{A_j}$} Finally, using \eqref{sm}, we can also reduce the cost of computing $M_{A_j}$, as:
\begin{align}\label{eq:matrix}
M_{A_j}=& \left[\begin{smallmatrix} M_A+\frac{\phi\phi^T}{1+v_j^T\Sigma_A^{-1}v_j}  & -\xi\\
-\xi^T & 1- v_j^T\xi\end{smallmatrix}\right]
\end{align}
where $\xi=V_A(\Sigma_{A_j}^{-1}v_j)$ and $\phi=V_A(\Sigma_{A}^{-1}v_j)$, which reduces the computation cost to $O(|A|^2+d^2)$ multiplications.
\medskip

In conclusion, using the above adaptive operations reduces the cost of computing $L_j$ by one order of magnitude to $O(|A|^2+d^2),$ which is optimal (as $M_A$ is an $|A|\times |A|$ matrix, and $\Sigma_A$ is $d\times d$). The rank-one adaptations yield such performance without sophisticated matrix inversion or multiplication algorithms, such as Strassen's algorithm. The we refer to resulting algorithm  as \incfbc;  we empirically
compare the two implementations in~\Cref{sec:eval}.

\subsection{Selection Through Point Estimation}
\label{sec:pointest}
An alternative method for selection can be constructed by replacing the exact estimation of the expected risk with a ``point estimate'' (see also \cite{silva2012active}). In fact, such a selection method can be easily combined with an arbitrary classifier that operates on user-provided ratings as input. This makes such an approach especially useful when the expected risk is hard to estimate in a closed form. We therefore outline this method below, noting however that several problems arise when the risk is computed through such a point estimation.

We describe the method for a general classifier $C$, also summarized in Algorithm \ref{algo:pe}.   Given a set of ratings $r_A$ over a set $A\subseteq\itemset$, the classifier $C$ returns  a probability
$\textstyle\Pr_C(t\mid r_A) ,$
for each type $t\in \typeset$. This is the probability that the user's type is $t$, conditioned on the observed ratings $r_A$.  Given a set of observed ratings $r_A$, we can estimate the type of the user using the classifier $C$ though maximum likelihood a-posteriori estimation, as
$\hat{t}(r_A) =\textstyle\argmax_{t\in \typeset} \Pr_C(t\mid r_A). $
Using this estimate, we can further estimate the most likely profile $\hat{u}\in \reals^d$ through ridge regression~\cite{hastie} over the observed ratings $r_A$ and the corresponding profiles $V_A$  (see Algorithm~\ref{algo:pe} for details). Using the estimated profile $\hat{u}$ and the estimated type $\hat{t}$, we can predict the rating of every item $j\in \itemset\setminus A$ as
$ \hat{r}_j  = \hat{u}^T v_j +z_{j\hat{t}}, $
and subsequently estimate the expected risk if the rating for item $j$ is revealed as
$ \textstyle\min_{t\in \typeset}\Pr_C(t\mid r_A\cup \hat{r}_j).$
We refer to this as a ``point estimate'', as it replaces the integration that the expected risk corresponds to with the value at a single point, namely, the predicted rating $\hat{r}_j$.

Using such estimates, selection can proceed as follows. Given the set of observed ratings $A$, we can estimate the risk of the classifier $C$ for every item $j$ in $\itemset\setminus A$ through the above estimation process, and pick the item with the minimum estimated risk. The rating of this item is subsequently revealed, and new estimates $\hat{t}$ and $\hat{u}$ can thusly be obtained, repeating the process until a desired confidence is reached.

\begin{algorithm}[!t]
  \caption{\pe \textsc{Active Learning}}\label{algo:pe}
{\fontsize{7}{7}\selectfont
  \begin{algorithmic}[1]
    \Require{ Item profiles $V$, item biases $Z$, classifier $C$, confidence $\tau$ }
    \State $A \leftarrow \emptyset$, $r_A \leftarrow \emptyset$
    \Repeat
    \State$\hat{\type} \leftarrow \argmax_{\type \in \typeset} \Pr_C(\type\mid r_A) $
    \State $\hat{u} \leftarrow (\lambda I+V_A^TV_A)^{-1}V_A^T(r_A-z_{A\hat{\type}})$
	\For{ \textbf{all} $j\in \itemset\setminus A$}
	 \State  $\hat{r}_j \leftarrow \hat{u}^T v_j +z_{j\hat{\type}}$
	 \State  $L_j \leftarrow \min_{\type\in\typeset}\Pr_C(t\mid r_{A}\cup \hat{r}_j)$
	\EndFor
         \State $j^* \leftarrow \arg\min_j L_j$
	     \State Query user to obtain $r_{j^*}$
         \State $A\leftarrow A\cup \{j^*\}$, $r_{A}\leftarrow r_A\cup r_{j^*}$
     \Until{$1-L_{j^*}>\tau$}
  \end{algorithmic}
}
\end{algorithm}

Clearly, point estimation avoids computing the expected risk exactly, which can be advantageous when the corresponding expectation under a given classifier $C$ can only be computed by numerical integration. This is not the case for \fbc, as we have seen, but this can be the only tractable option  for an arbitrary classifier.  Unfortunately, this estimation can be quite inaccurate in practice, consequently leading to poor performance in selections; we observe such a performance degradation in our evaluations (\Cref{sec:eval}). Put differently, a point estimate of the risk takes into account what the predicted rating of an item $j$ is in \emph{expectation}, and how this rating can potentially affect the risk; however, it does not account for how \emph{variable} this prediction is. A highly variable prediction might have a very different expected risk; the exact computation of the expectation does take this into account whereas point estimation does not.

\makeatletter{}

\section{Evaluation}
\label{sec:eval}

In this section we evaluate the performance of our methods using real datasets. We begin by describing the datasets and experiments, then perform a comparative analysis of  both passive  and active methods.

\ignore{
\begin{table*}[!t]
\small
\begin{center}
\begin{tabular}{|l|c|c|c|c|c|c|}
\hline
  &  \multicolumn{3}{|c|}{Class 1 (Males / Young adults)} & \multicolumn{3}{|c|}{Class 2 (Females / Adults)} \\
\hline
  & Users & Ratings & Max Ratings & Users & Ratings & Max Ratings \\
\hline
Gender   & 4319 &  750045 & 2150 & 1703 & 245109 & 1265\\
\hline
Age   & 2619 &  391436 & 2150 & 3416 & 603718 & 1801\\
\hline
\end{tabular}
\caption{Statistics of the evaluated datasets, consisting of 6K users, 3K items and almost 1M ratings.}
\label{table:stats}
\vspace{-5mm}
\end{center}
\end{table*}
\begin{table*}[t]
\small
\begin{center}
\begin{tabular}{|l|c|c|c|c|c|c|c|c|}
\cline{2-9}
\multicolumn{1}{l|}{} & \multicolumn{3}{|c|}{\ml} & \multicolumn{3}{|c|}{Politics-and-TV} & \multicolumn{2}{|c|}{Flixster}\\
\cline{2-9}
\multicolumn{1}{l|}{} & Total & Gender & Age & Total & Gender & Political Views & Total & Gender \\
\hline
Users & 6K & 2.5:1 & 1:1.3 & 992 & 1:1.8 & 1:1.6 & 26K & 1.7:1 \\
\hline
Items & 3K & - & - & 50 & - & - & 9921 & - \\
\hline
Ratings & 1M & 3:1 & 1:1.6 & 29.9K & 1:2.1 & 1:1.5 & 5.6M & 1.5:1  \\
\hline
\end{tabular}
\caption{Datasets statistics of the evaluated datasets.}
\label{table:stats}
\end{center}
\vspace{-5mm}
\end{table*}
}

\begin{table}[t!]
{\fontsize{7}{7}\selectfont
\begin{center}
\begin{tabular}{|l|l|c|c|c|}
\hline
Dataset& Type & Users & Items & Ratings \\
\hline
  & All & 6K & 3K & 1M \\
\ml & Gender (Female:Male) & 1:2.5 & - & 1:3 \\
& Age (Young:Adult) & 1:1.3 & -  &1:1.6 \\
\hline
 & All & 992 & 50 & 29.9K \\
PTV & Gender (Female:Male) & 1.8:1 & - & 1.6:1 \\
& Political Views (R:D) & 1:1.6 & - & 1:2.1 \\
\hline
 & All & 26K & 9921 & 5.6M \\
\fl & Gender (Female:Male) & 1.7:1 & - & 1.5:1 \\
\hline
\end{tabular}
\caption{Dataset statistics.}
\label{table:stats}
\end{center}
}
\vspace{-5mm}
\end{table}

\techrep{
\floatsetup[figure]{style=plain,subcapbesideposition=center}
\begin{figure}[t!]
\centering
\sidesubfloat[\ml]{
\includegraphics[width=0.7\textwidth]{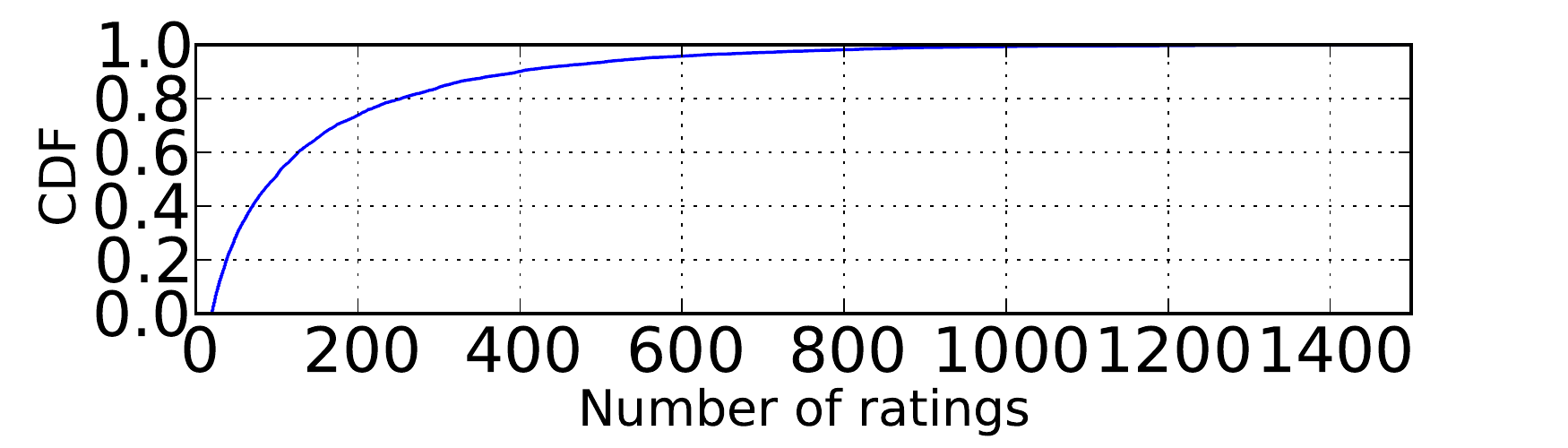}
}\\
\sidesubfloat[Flixster]{
\includegraphics[width=0.7\textwidth]{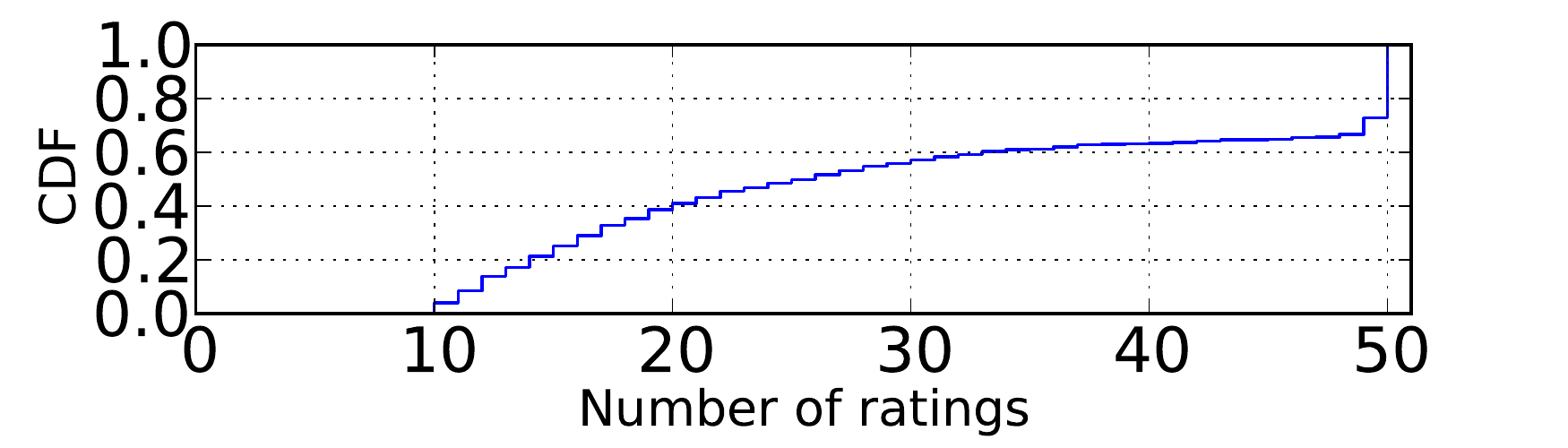}
}\\
\sidesubfloat[PTV]{
\includegraphics[width=0.7\textwidth]{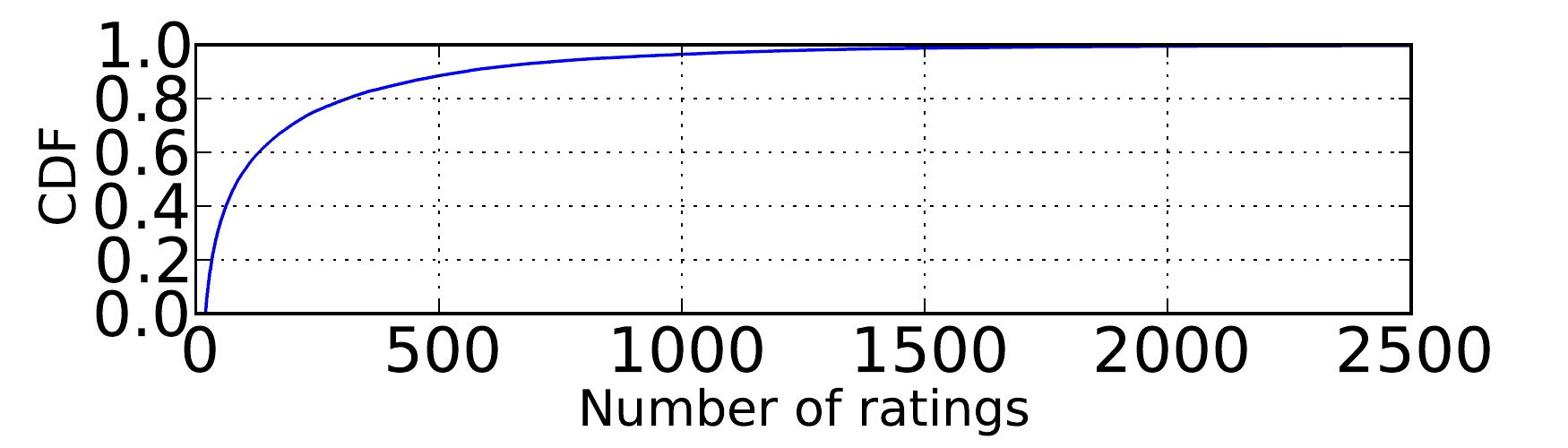}
}
\caption{Cumulative distribution of number of ratings per user for (a) \ml (b) Politics-and-TV (c) \fl.}
\label{fig:ratings}
\end{figure}
}{}

\subsection{Experimental Setup}
\para{Datasets} 
We evaluate our method using three datasets:  \ml, Flixster~\cite{Jamali:2010}, and Politics-and-TV (PTV)~\cite{salman:2013}. The \ml dataset includes users' ratings for movies alongside with the users' gender and age. For simplicity, we categorize the age group of users as \emph{young adults} (ages 18--35), or \emph{adults} (ages 36--65). Flixster is a similar movie ratings   dataset, and contains user gender information. PTV includes ratings by US users on 50 different TV-shows, along with each user's gender and political affiliation (Democrat or Republican).
 We preprocessed \ml and Flixster to consider only users with at least 20 ratings,
and items that were rated by at least 20 users. Since PTV includes only 50 TV-shows, we preprocessed the data
to ensure that each user has at least 10 ratings. \Cref{table:stats} summarizes the datasets used for evaluation.
For each user type, the table shows the ratio between the number of users of one type versus the other type (as labeled in the table). \techrep{\Cref{fig:ratings} shows the cumulative distribution function (CDF) of the number of ratings per user
across the three preprocessed datasets. We see that for the \ml and \fl datasets, there are many users with hundreds of items rated in their profile.
}{Further details on the datasets can be found in \cite{techrep}.}

\para{Evaluation Method} 
In our setting, the recommender system infers user attributes 
from a set of strategically selected items.
To understand the effectiveness of \fbc compared to other classification methods in an adversarial setting, we perform the following evaluation.
We first split the dataset into a training set (\eg, users that are willing to disclose 
the private attribute) and evaluation set (\eg, users that are privacy-sensitive), 
and train different classifiers on the training set -- \eg, in the case of \fbc we learn the item profiles and biases. 
Then, for each user in the evaluation set, we incrementally select items for the user to rate. In the passive methods, the selection of next item does not depend on the previous ratings provided by the user, whereas 
in the active methods it does.

After the user rates an item, we use the classifier to infer 
the private type. For any user, since we only have the rating information for the set of movies 
that she has rated, we limit the selection process to this set. 
Users may have rated different number of movies, for instance, 
roughly 50\% of the users of \ml have rated less than 100 movies out of 3000 
\techrep{(see \Cref{fig:ratings}).}{(see \cite{techrep}).}
Therefore, we limit the number of questions asked to 100 for \ml and \fl and all 50 for \ptv.
Unless specified, all evaluations of \fbc were done
using the efficient incremental implementation \incfbc. 

\para{Evaluation Metrics}
We evaluate   classification performance through the area under the curve (AUC) metric, 
and prediction performance through the root mean squared error (RMSE). 
If a recommender system uses our method to maliciously learn user features,
it is important that such a mechanism for strategic solicitation of ratings has a minimal
impact on the quality of recommendations, otherwise the user may detect its hidden agenda.
We measure the quality of recommendations by holding out an evaluation set of
10 items for each user.  After every 10 questions (solicited ratings) we
predict the ratings on the evaluation set by applying ridge regression using the provided
ratings and item profiles to learn a user profile. We predict the ratings on the evaluation set and compute the RMSE over all users.

\para{Parameter settings}
We split each dataset into training and testing and perform MF with 10-fold cross validation.
We learn the item latent factors required by \fbc using the training set, with type biases
for age, gender and political affiliation as applicable to the three datasets. For MF, we  use 20 iterations
of stochastic gradient descent~\cite{Koren:2009} to minimize \eqref{mle}, using the same regularization parameter for users and movies. Through 10-fold cross validation we determined the optimal dimension to be $d=20$, and the optimal regularization parameter to be $0.1$, for each of the biases.
We also compute the optimal $\lambda$ used in the classifier \eqref{fbc} through 10-fold cross validation to maximize the AUC, resulting in
$\lambda = 100$ for gender and $\lambda = 200$ for age for the \ml dataset, $\lambda = 10$ for gender and political views for the \ptv dataset, and $\lambda = 200$ for gender for the \fl dataset.

\begin{figure}[t]
\centering
\includegraphics[width=\figtwo]{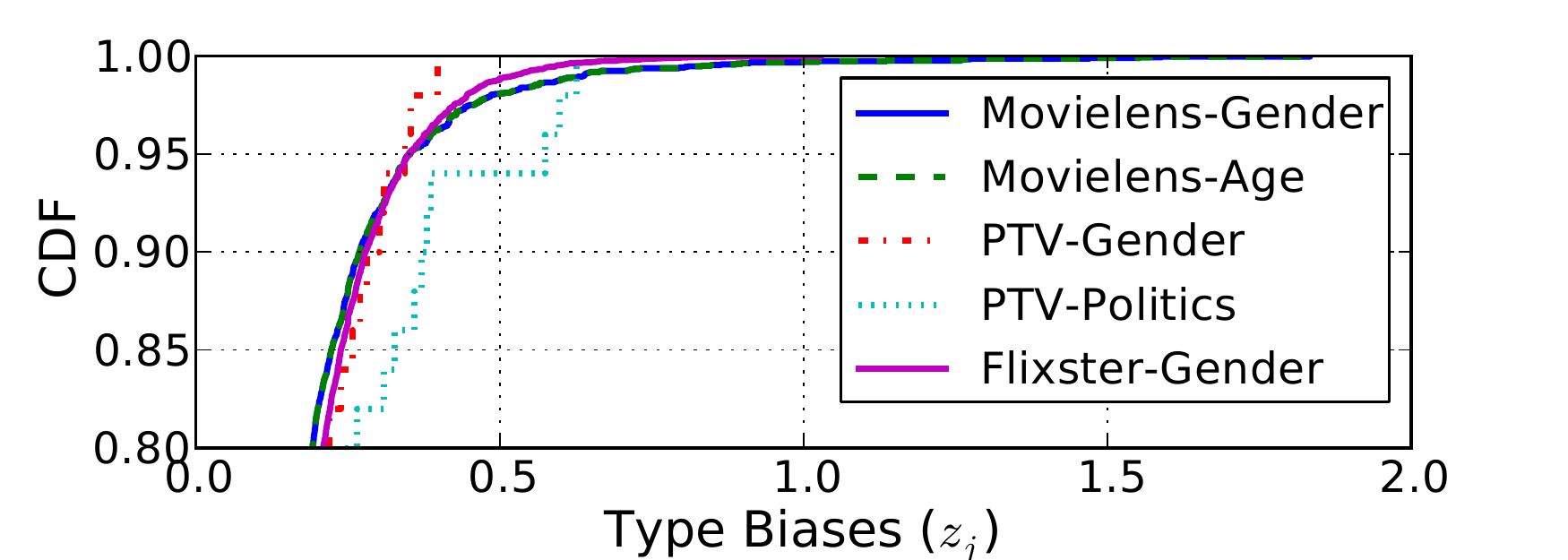}
\caption{Cumulative distributions of the type bias ($|z_j|$) for the different datasets, zoomed to the top 20\% of items.}
\label{fig:gaps}
\vspace{-6pt}
\end{figure}

\begin{figure*}[t!]
\subfloat[AUC with passive learning]{
\includegraphics[width=\textwidth]{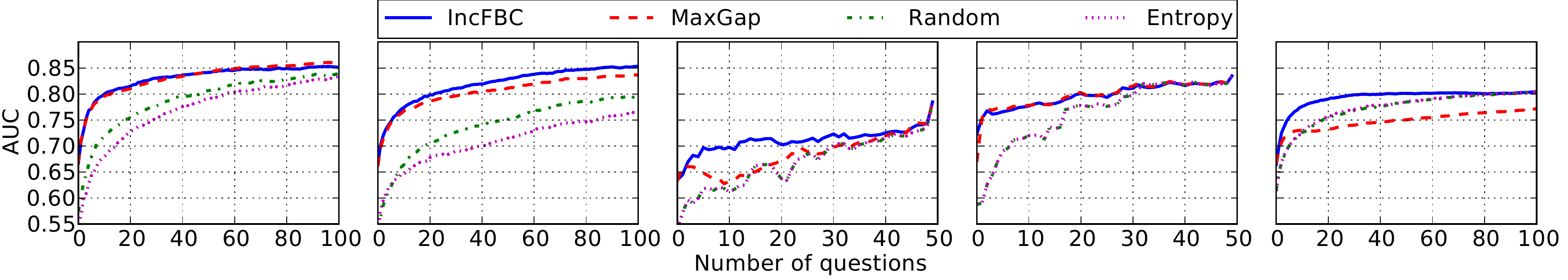}
\label{fig:auc_passive} } \\
\subfloat[RMSE with passive learning]{
\includegraphics[width=\textwidth]{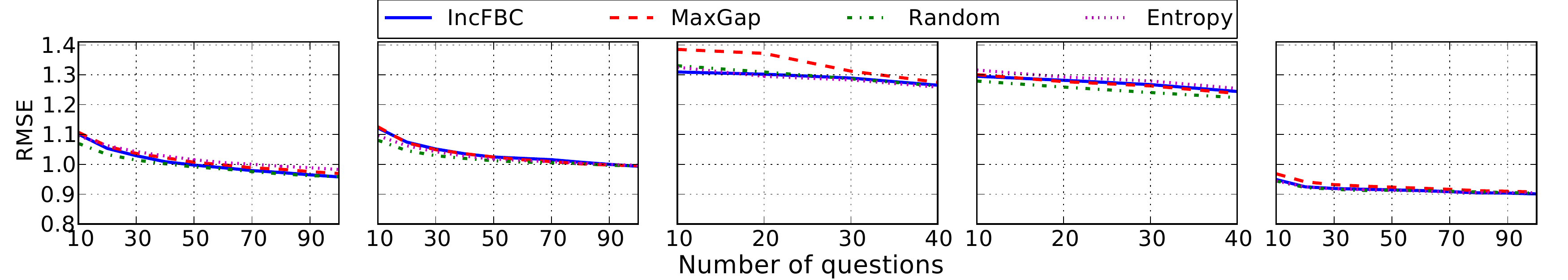}
\label{fig:rmse_passive} }
\caption{Average AUC and RMSE of the \fbc classifier with increasing questions for different passive selection strategies, and \incfbc for comparison. Datasets (left to right) -- \ml-Gender, \ml-Age, \ptv-Gender, \ptv-Political Views, \fl-Gender.}
\label{fig:passive}
\vspace{-6pt}
\end{figure*}

\subsection{Passive Learning}
\Cref{fig:passive} shows the AUC and RMSE obtained using \maxgap, and two other passive methods -- Random and Entropy. 
For reference,  we also show the performance of our active method \incfbc. 
Random selection is a natural baseline as users may rate items in any
arbitrary order. The second method, Entropy, presents items to the user in descending order of their rating entropy, \ie, start with items that have polarized ratings. This method was shown to
be efficient in a cold-start setting in~\cite{Rashid02gettingto} as it can quickly build
user profiles in a matrix factorization based recommender system.

\para{AUC}
\Cref{fig:auc_passive} shows that \maxgap performs significantly better than the other passive methods on both \ml and \ptv datasets. For the first 10 questions, which are critical when considering the need for quick inference, it is the best passive method on all datasets. 
Interestingly, on \ml-Gender and \ptv-Politics, \maxgap performs very similar to the adaptive, more complex \incfbc. As a result of the greedy nature of \maxgap, we expect it to perform well on datasets that have items with large biases. 
To better understand \maxgap's performance, \Cref{fig:gaps} shows the top 20\% of the cumulative distributions of the type biases over the set of items in the different datasets. The plot clearly shows that the items in \ptv-Gender have the lowest biases, resulting in the poorest performance of \maxgap. Conversely, in \ml-Gender, the biases are the largest, thus \maxgap performs well, in par with the adaptive \incfbc. In \ptv-Politics the biases are not overly high, but since most users rate all items, a few discriminating items are sufficient to enable \maxgap to perform well. This observation is supported by the findings of~\cite{Calmon:2012} that identified 5 TV-shows that immediately reveal a user's political views.

\para{RMSE}
\Cref{fig:rmse_passive} plots the RMSE over increasing number of rated items, for \maxgap, Random, and Entropy, along with \incfbc. Even though \incfbc and \maxgap are designed to explore a specific attribute of the user's profile, they perform very well. Their RMSE is very close to that of Random and Entropy, with the \maxgap visibly worse only in one case, the \ptv-Gender dataset. 
 Since \incfbc and MaxBias focus on quickly learning a singe attribute of the user's profile, it is expected that they perform worse than the other methods, that aim to explore attributes more broadly. 
However, the figures show that \incfbc and \maxgap are almost identical to the other methods, and \maxgap only perform worse in the \ptv-Gender dataset. 
Moreover, in all datasets, \incfbc performs close to a random selection, indicating that \incfbc does not incur significant impact on the RMSE relative to an arbitrary order in which a user may rate items. Finally, \incfbc has an RMSE similar to the entropy method, which is designed to improve the RMSE in a cold-start setting.
 
These results show that a malicious recommender system that uses \incfbc to infer a private attribute of its users can also use the solicited ratings to provide recommendations, making it difficult for users to detect the hidden agenda.

\begin{figure*}[t!]
\subfloat[AUC with active learning]{
\includegraphics[width=\textwidth]{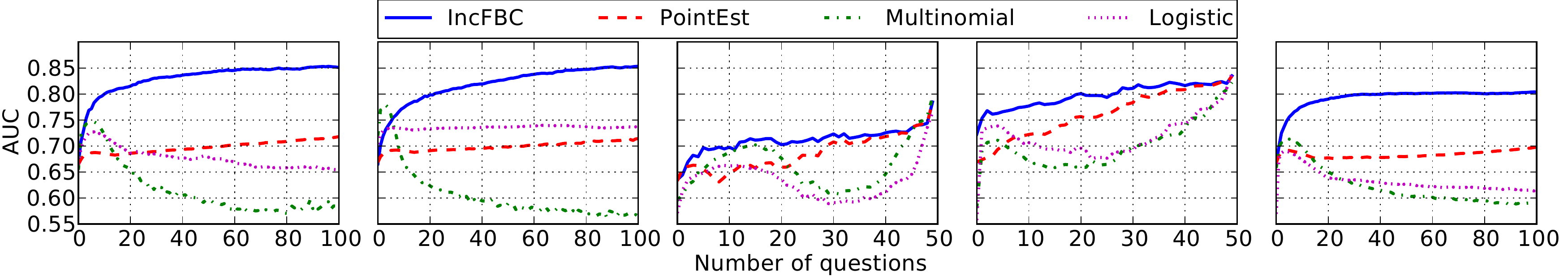}
\label{fig:auc_active}
}  \\
\subfloat[RMSE with active learning]{
\includegraphics[width=\textwidth]{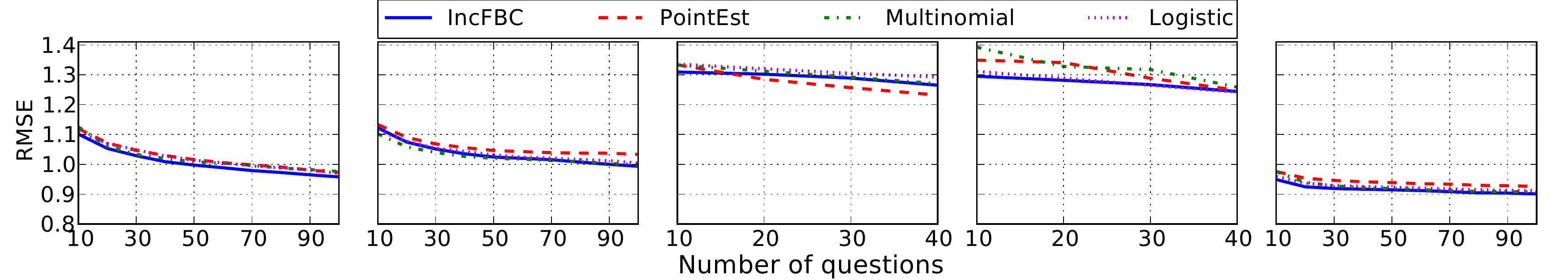}
\label{fig:rmse_active}
}
\caption{Average AUC and RMSE per number of questions for the three classifiers: \incfbc, multinomial (using \pe selector) and logistic (using \pe selector). Datasets from left to right -- \ml-Gender, \ml-Age, \ptv-Gender, \ptv-Political Views, \fl-Gender.}
\label{fig:classifiers}\label{fig:active}
\vspace{-3mm}
\end{figure*}

\subsection{Active Learning}

We compare our selection method to the logistic and multinomial classifiers by adapting them to an active learning setting. These classifiers were the top performing among those studied in previous work~\cite{blurme:2012} for gender prediction. Following~\cite{blurme:2012}, we train both of these classifiers over rating vectors padded with zeros: an item not rated by a user is marked with a rating value of $0$. 
In order to use logistic and multinomial classifiers in an active learning setting we use the point-estimate (\pe) method as described in \Cref{sec:pointest} (see Algorithm \ref{algo:pe}).
For the remainder of this section we refer to \pe with a logistic and multinomial classifiers as \emph{Logistic} and \emph{Multinomial}, respectively.

\sloppy
\para{AUC}
\Cref{fig:auc_active} plots the AUC of classification for a given
number of question using \pe and \incfbc selection, for all datasets.
\pe selector enables us to compare \fbc with the other classifiers for which 
we do not have a closed-form solution for selection. 
In all datasets, the plots show that \incfbc significantly outperforms both logistic and multinomial within a few questions, and reaches an improvement in AUC of 10--30\% in the \ml and \fl datasets.
\ignore{
Using \incfbc with just 20 movies per user we obtain a classification accuracy that is close to that obtained by static inference techniques which use the complete dataset. } \pe with the other classifiers is unable to achieve a good classification accuracy. \ignore{
\pe with the other two classifiers starts in some cases with a relatively good accuracy (around 0.7), however, these are essentially the class priors, as can be seen in~\Cref{table:stats}. Furthermore, 
the improvement in accuracy is extremely slow, indicating that these classifiers do not substantially improve beyond the class priors, whereas \incfbc exhibits a much faster increase in accuracy.
}

To put this in perspective, \Cref{table:classification} shows the performance of these classifiers, and that of a non-linear classifier SVM with RBF kernel, using all user ratings.  Note that this table considers \emph{all} ratings performed by \emph{all} users in each dataset, 
whereas the plots  in \Cref{fig:auc_active} show the average AUC computed over the 
users that have rated the indicated number of questions.
Logistic 
and in some cases multinomial classifiers perform significantly better than \fbc,
when classifying over the entire dataset. 
This shows that although any of these classifiers could be used for a static attack~\cite{blurme:2012}, \fbc is better suited to adaptive attacks with fewer available ratings. For instance, using \incfbc with just 20 movies per user we obtain a classification accuracy that is reasonably close to that obtained by static inference techniques which use the complete dataset. 

\para{RMSE}
For completeness, \Cref{fig:rmse_active} provides the RMSE using the different active methods, showing that \incfbc has a lower RMSE on almost all 
datasets. 

\ignore{
\Cref{fig:confidence} shows the average corresponding classification confidence. Compared to \Cref{fig:accuracy}, we see that both logistic and multinomial are ``over-confident'', in that they assign high confidence to users even while failing to classify them correctly. Moreover, in the PTV dataset, the final TV-shows selected by both logistic and multinomial actually reduce their confidence. These are TV-shows that the selector would not ask to rate in the presence of better options, but are eventually selected because of the limited number of shows available (50). In contrast, \incfbc is consistent and robust to the addition of such items, increasing both in terms of confidence and accuracy.
}

\ignore{
\begin{figure}[t]
\centering
\includegraphics[width=\figtwo]{fig/accuracy_selectors_lambda}
\caption {Effect of $\lambda$ on \pe in Movielens-Gender.}
\label{fig:lambda}
\vspace{-6pt}
\end{figure}
}

\begin{table}[t]
\small
\begin{center}
\begin{tabular}{|l|c|c|c|c|c|}
\cline{2-6}
\multicolumn{1}{c|}{} & \multicolumn{2}{|c|}{\ml} & \multicolumn{2}{|c|}{\ptv} & Flixster\\
\cline{2-6}
\multicolumn{1}{c|}{} & Gender & Age & Gender & Politics & Gender \\
\hline
\fbc             & 0.827 & 0.825 & 0.683 & 0.748 & 0.650 \\
\hline
Logistic       	 & 0.865 & 0.906 & 0.756 & 0.778 & 0.861 \\
\hline
Multinomial 	 & 0.810 & 0.817 & 0.753 & 0.758 & 0.747 \\
\hline
SVM (RBF)  & 0.838 & 0.893 & 0.613 &  0.737 & NA  \\
\hline
\end{tabular}
\caption{AUC of classification with full user history.}
\label{table:classification}
\end{center}
\vspace{-5mm}
\end{table}

\begin{figure}[t]
\centering
\includegraphics[width=\figtwo]{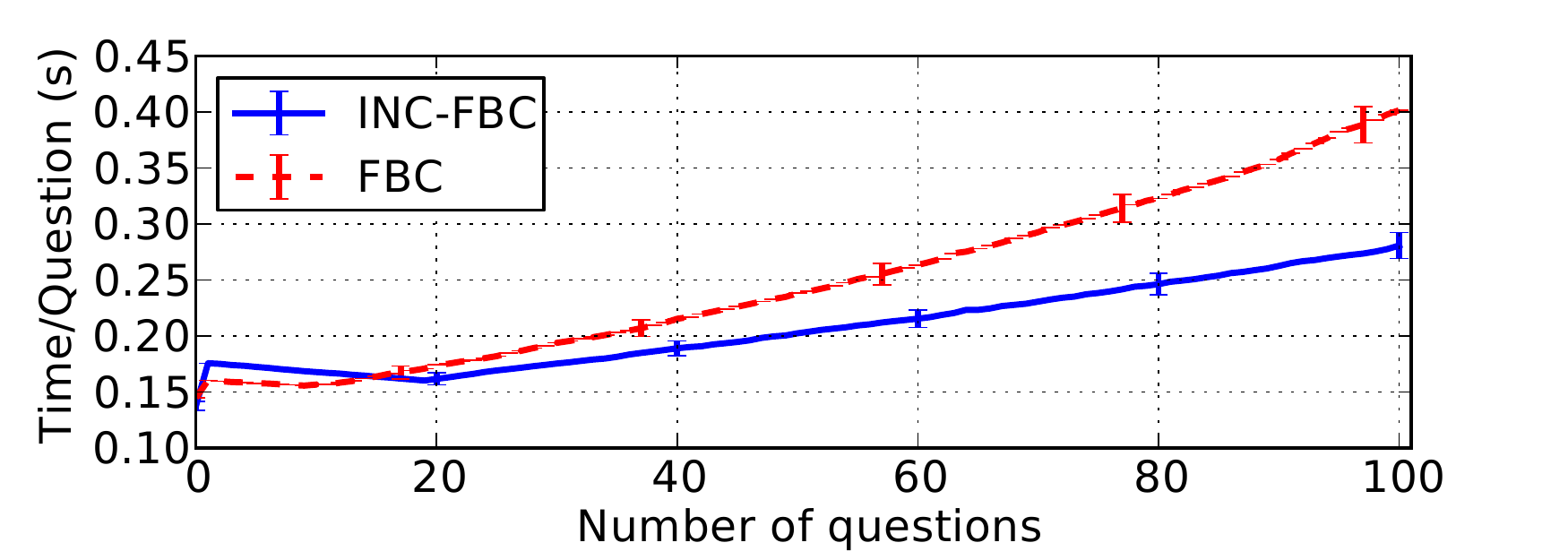}
\caption{Running time improvement of \incfbc over \fbc.}
\label{fig:time}
\vspace{-6pt}
\end{figure}

\para{Running Time} Finally, we seek to quantify the improvement in execution time obtained by the incremental computations of \incfbc. We ran both \fbc and \incfbc on a commodity server with
a RAM size of 128GB  and a CPU speed of 2.6GHz.
\Cref{fig:time} shows the average time per movie selection for both \fbc and \incfbc for increasing
number of questions (movies presented to the user). The error bars depict the 95\% confidence interval
surrounding the mean. The plot shows that when the number of questions is small the time
per question is relatively constant, and increases with the number of questions. 
As discussed in \Cref{efficientimpl}, when the number
of questions is smaller than the dimension of the factor vectors (in our case $d=20$),
the complexity of the efficient algorithm is dominated by $d$.
In the first few questions \fbc is slightly
faster than \incfbc as a result of the efficient implementation of inversion for small matrices. 
However, as the matrix becomes larger, the size of the matrix dominates the complexity and the
incremental computations performed in \incfbc are significantly faster than \fbc, reaching
a speedup of 30\%.

\makeatletter{}\section{Conclusion and Future Work}
\label{sec:concl}

We presented a new attack that a recommender system could use to pursue a hidden agenda of inferring private attributes for users that do not voluntarily disclose them. Our solution, that includes a mechanism to select which question to ask a user, as well as a classifier, is efficient both in terms of the number of questions asked, and the runtime to generate each question.  Moving beyond binary attributes to multi-category attributes\techrep{, using the relationship of our classifier to LDA,}{} is an interesting open question.  Exploring the attack from the user's perspective, to better advise users on ways to identify and potentially mitigate such attacks is also an important future direction.

\ignore{
In this paper we present a new effective attack that a recommender system can employ in order to quickly learn private attributes of its users. The attack leverages the widely used matrix factorization in order to infer, in an active learning setting, user social and demographic details, based solely on
ratings given to the presented items. We show that using our method, the system can obtain a classification accuracy
that is comparable or even better than the one obtained from the entire dataset, by using only
a fraction of the ratings.}

\ignore{
By leveraging the techniques presented in this work, recommender systems that
require demographic information for improved personalization
can quickly and accurately learn their users' demographic details, while incurring minimal impact on
the quality of their recommendations.
}
\ignore{
We plan to pursue several directions in future work. First, we plan to extend the attack into multi-class attributes. Second, we plan to study the problem of a system trying to simultaneously infer multiple private attributes, \eg, age and gender. Third, we wish to evaluate the attack when the user has already rated several items, and the recommender system seeks to improve its confidence in the inferred attribute. Finally, we seek to explore the attack from the user's perspective, so that we can better advise users on ways to identify and potentially mitigate such attacks.}

\bibliographystyle{abbrv}

\appendix
\makeatletter{}

\para{Derivation of Equation~{\protect\lowercase{\eqref{eqn:risk}}} }\label{app:derive}
For $A_j=A\cup\{j\}$ and $\type^c$ the binary complement of $\type$, the expected risk $\expect[L(\hat{\type}(r_{A_j}))\mid r_A]$, if the rating for movie $j$ is revealed is
\begin{align*}& \int_{r_j\in \reals} \Pr(\hat{t}^c(r_{A_j})\mid r_{A_j}) \Pr(r_{A_j}\mid r_A) dr_j\\
& =  \int_{r_j\in \reals}  \Pr(\bar{b}(r_{A_j})\mid r_{A_j}) \frac{\Pr(r_{A_j})}{\Pr(r_A)} dr_j\\
&\stackrel{\eqref{uncond},\eqref{gender}}{=}C\frac{\displaystyle\int_{r_j\in\reals}  e^{y_{\hat{\type}^c(r_{A_j})}^T\left(V_{A_j}\Sigma_{A_j}^{-1}V_{A_j}^T-I\right)y_{\hat{b}^c(r_{A_j})}/2\sigma_0^2}dr_j
}{\sqrt{\det(\Sigma_{A_j})}}\end{align*}
where $y_{\hat{\type}(r_{A_j})} ={r}_{A_j}- z_{A_j \hat{\type}^c(r_{A_j})} $,  $\Sigma_{A_j} = \lambda I + V_{A_j}^TV_{A_j},$
and $C$ a term that does not depend on $j$. The expected risk is thus proportional to:
$$L_j = \frac{\int_{r_j\in\reals} e^{-({r}_{A_j}-\bias_{A_j \hat{t}^c(r_{A_j})})^TM_{A_j}({r}_{A_j}-\bias_{A_j \hat{t}^c(r_{A_j})})/2\sigma_1^2}dr_j
}{\sqrt{|\Sigma_{A_j}|}}  $$
where $|\Sigma_{A_j}| = \det(\Sigma_{A_j})$ $M_{A_j}=I-V_{A_j}\Sigma_{A_j}^{-1}V_{A_j}^T$ and $\hat{\type}^c(r_{A_j})$ the complement of prediction under $r_{A_j}$. Under Theorem~\ref{thm:fbc}, as $\hat\type$ is given by \eqref{fbc}, $L_j$ simplifies to \eqref{eqn:risk}.\hspace*{\stretch{1}}\qed

\para{Closed Form of {\protect\lowercase{\eqref{eqn:risk}}}}\label{app:closed}
Let $\xi=V_A(\Sigma_{A_j}^{-1}v_j)$, $\phi=V_A(\Sigma_{A}^{-1}v_j)$, $\mu_1 = M_A+\frac{\phi\phi^T}{1+v_j^T\Sigma_A^{-1}v_j}$ and 
$\mu_2= 1- v_j^T\Sigma_{A_j}^{-1}v_j$. Then, from \eqref{eq:matrix} we get that:
\begin{align*}
\bar{r}_{A_j}^T &M_{A_j}\bar{r}_{A_j}+ 2|\diff_{A_j}^TM_{A_j} \bar{r}_{A_j}|+\diff_{A_j}^T M_{A_j}\diff_{A_j}=\\
&	   \mu_2\bar{r}_j^2 -2\bar{r}_A^T\xi \bar{r}_j + 
	    |(\delta_j\mu_2-\delta_A^T\xi )\bar{r}_j +\delta_A^T\mu_1\bar{r}_A-\delta_j\xi^T\bar{r}_A|\\
             &\qquad\qquad +   \bar{r}_A^T\mu_1\bar{r}_A +\delta_{A_j}^T M_{A_j}\delta_{A_j}
\end{align*}
For simplicity of exposition, let 
$\alpha_1=\mu_2/\sigma_1^2, 
\alpha_2= -2r_A^T\xi/\sigma_1^2 , 
\alpha_3=(z_j\mu_2-z_A^T\xi)/\sigma_1^2, 
\alpha_4=(z_A^T\mu_1-z_j\xi^T)r_A/\sigma_1^2, 
\alpha_5=(r_A^T\mu_1r_A +z_{A_j}^T M_{A_j}z_{A_j})/\sigma_1^2  $
If $\alpha_3>0$, 
substituting these in the risk, we have,
\begin{align*}
L_j 	= \frac{1}{\sqrt{|\Sigma_{A_j}|}}&\Big(\int_{r_j=-\infty}^{-\frac{\alpha_4}{\alpha_3}} e^{-(
	  \alpha_1r_j^2 +(\alpha_2 -\alpha_3)r_j -\alpha_4+ \alpha_5 )/2}dr_j\\
	    &+\int_{r_j=-\frac{\alpha_4}{\alpha_3}}^{\infty} e^{-(
	  \alpha_1r_j^2 +(\alpha_2 +\alpha_3)r_j +\alpha_4+ \alpha_5)/2}dr_j\Big)
\end{align*}
Letting $x=\sqrt{\alpha_1}r_j+ \frac{\alpha_2-\alpha_3}{2\sqrt{\alpha_1}}$
and $y=\sqrt{\alpha_1}r_j+ \frac{\alpha_2+\alpha_3}{2\sqrt{\alpha_1}}$, 
and substituting $dx=dy=\sqrt{\alpha_1}dr_j$
we can rewrite the above as,
\begin{align*}
L_j  =& 
           \frac{1}{\sqrt{\alpha_1|\Sigma_{A_j}|}}\Big(e^{\frac{\frac{(\alpha_2-\alpha_3)^2}{4\alpha_1}-\alpha_5+\alpha_4}{2}}   h\left(\frac{\alpha_2-\alpha_3}{2\sqrt{\alpha_1}}-\frac{\alpha_4\sqrt{\alpha_1}}{\alpha_3}\right)+\\&e^{\frac{\frac{(\alpha_2+\alpha_3)^2}{4\alpha_1}-\alpha_5-\alpha_4}{2}}  h\left(-\frac{\alpha_2+\alpha_3}{2\sqrt{\alpha_1}}+\frac{\alpha_4\sqrt{\alpha_1}}{\alpha_3}\right)\Big)   
\end{align*}
where $h(x)=\int_{-\infty}^x e^{-x^2/2}dx$, which can be expressed in terms of the error  function ($\mathrm{erf}$). A similar derivation applies if $\alpha_3\leq 0$.
\hspace*{\stretch{1}}\qed

\end{document}